\documentclass[10pt,twocolumn,letterpaper]{article}

\usepackage[final]{cvpr}

\usepackage{times}
\usepackage{epsfig}
\usepackage{graphicx}
\usepackage{amsmath}
\usepackage{amssymb}

\usepackage[utf8]{inputenc}
\usepackage{multirow}
\usepackage{rotating}
\usepackage{verbatim}
\usepackage{booktabs}
\usepackage{array}
\usepackage{textcomp, gensymb}
\usepackage{diagbox}
\usepackage[T1]{fontenc}    
\usepackage{url}            
\usepackage{booktabs}       
\usepackage{amsfonts}       
\usepackage{nicefrac}       
\usepackage{microtype}      
\usepackage{setspace}
\usepackage{multirow}
\usepackage{amssymb}
\usepackage{diagbox}
\usepackage{mathtools}
\usepackage{wrapfig}
\usepackage{diagbox} 
\usepackage{wrapfig,lipsum,booktabs}
\usepackage{caption}
\usepackage{appendix}

\usepackage{color}
\usepackage{xcolor}
\definecolor{citecolor}{rgb}{0.21,0.49,0.74}
\usepackage{colortbl}
\usepackage{array}

\usepackage[pagebackref=false,breaklinks=true,citecolor=citecolor,colorlinks,bookmarks=false]{hyperref}

\definecolor{tblue}{RGB}{80,80,245}
\definecolor{tred}{RGB}{250,100,100}

\newcolumntype{x}[1]{>{\centering\arraybackslash}p{#1pt}}

\newcommand{\app}{\raise.17ex\hbox{$\scriptstyle\sim$}}

\newlength\savewidth\newcommand\shline{\noalign{\global\savewidth\arrayrulewidth
  \global\arrayrulewidth 1pt}\hline\noalign{\global\arrayrulewidth\savewidth}}
\newcommand{\tablestyle}[2]{\setlength{\tabcolsep}{#1}\renewcommand{\arraystretch}{#2}\centering\footnotesize}

\usepackage{amssymb}
\usepackage{pifont}

\usepackage[pagebackref=false,breaklinks=true,citecolor=citecolor,colorlinks,bookmarks=false]{hyperref}

\usepackage{amssymb}
\usepackage{pifont}

\definecolor{tgreen}{RGB}{32,178,170}

\definecolor{tgray}{RGB}{169,169,169}

\definecolor{tbg}{RGB}{230,245,230}
\definecolor{tbb}{RGB}{135,206,250}
\definecolor{blue2}{RGB}{0,139,139}
\definecolor{red2}{RGB}{255,127,80}

\newcommand{\cmark}{\color{tgreen}\ding{51}}%
\newcommand{\xmark}{\color{red2}\ding{55}}%

\newcommand{\dataname}{MegaSynth}

\def\paperID{2866}
\def\confName{CVPR}
\def\confYear{2025}

\title{\dataname{}: Scaling Up 3D Scene Reconstruction with Synthesized Data}

\author{
Hanwen Jiang$^{1}$~~~
Zexiang Xu$^{2}$~~~
Desai Xie$^{3}$~~~
Ziwen Chen$^{4}$~~~
Haian Jin$^{5}$~~~
Fujun Luan$^{2}$~~~
Zhixin Shu$^{2}$\\
Kai Zhang$^{2}$~~~
Sai Bi$^{2}$~~~
Xin Sun$^{2}$~~~
Jiuxiang Gu$^{2}$~~~
Qixing Huang$^{1}$~~~
Georgios Pavlakos$^{1}$~~~
Hao Tan$^{2}$~~~
\\
$^{1}$The University of Texas at Austin \ 
$^{2}$Adobe Research \ \\
$^{3}$Stony Brook University \ 
${^4}$Oregon State University \ 
${^5}$Cornell University \ \\
\small{Project \& Code: \href{https://hwjiang1510.github.io/MegaSynth/}{https://hwjiang1510.github.io/MegaSynth/}}
}

\begin{document}
\twocolumn[\maketitle\vspace{0mm}\begin{center}
    \captionsetup{type=figure}
    \vspace{-0.35in}
    \includegraphics[width=\textwidth]{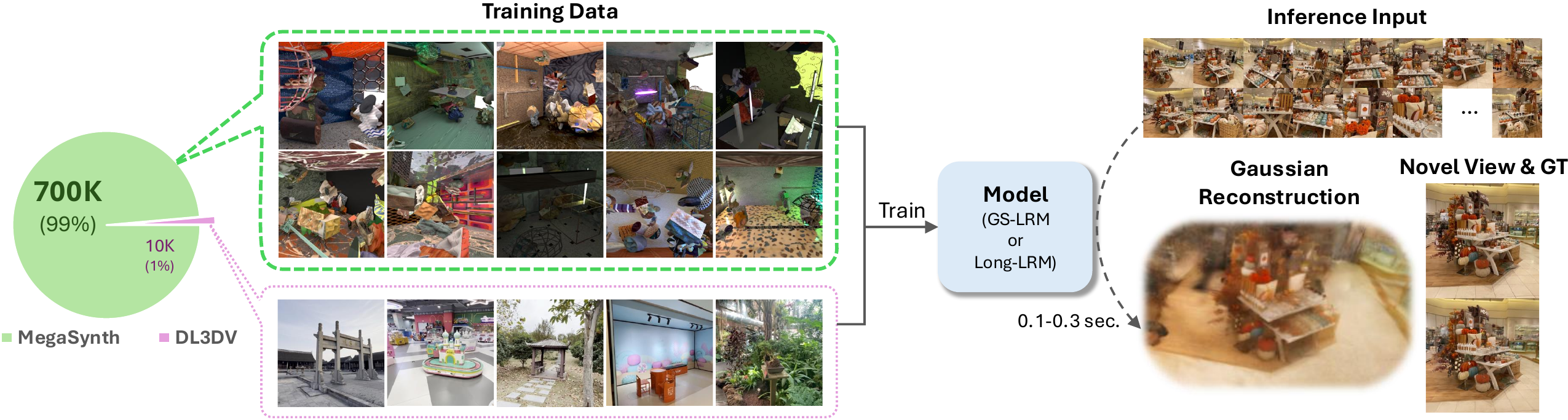}
    \vspace{-0.27in}
    \captionof{figure}{%
        \small{We introduce \dataname{}, a \textbf{non-semantic} synthesized dataset for training LRMs. \dataname{} benefits from its scalability and controllability, enabling us to generate 700K scenes in 3 days. We train LRMs with both the large-scale \dataname{} data and small-scale real data, improving LRMs for reconstructing wide-coverage scenes from dense-view images. }  
    }
    \label{fig: teaser}
    \vspace{0.05in}
\end{center}\bigbreak]

\begin{abstract}
   We propose scaling up 3D scene reconstruction by training with synthesized data. At the core of our work is \dataname{}, a procedurally generated 3D dataset comprising 700K scenes—over 50 times larger than the prior real dataset DL3DV—dramatically scaling the training data. To enable scalable data generation, our key idea is eliminating semantic information, removing the need to model complex semantic priors such as object affordances and scene composition. Instead, we model scenes with basic spatial structures and geometry primitives, offering scalability.
   Besides, we control data complexity to facilitate training while loosely aligning it with real-world data distribution to benefit real-world generalization. 
   We explore training LRMs with both \dataname{} and available real data.
   Experiment results show that joint training or pre-training with \dataname{} improves reconstruction quality by 1.2 to 1.8 dB PSNR across diverse image domains. Moreover, models trained solely on \dataname{} perform comparably to those trained on real data, underscoring the low-level nature of 3D reconstruction. Additionally, we provide an in-depth analysis of \dataname{}'s properties for enhancing model capability, training stability, and generalization, as well as application to other tasks. 
\end{abstract}

\section{Introduction}

The scaling law has shifted the focus of contemporary AI research toward large foundation models, which are built with scalable neural network architectures~\cite{vaswani2017transformer, ho2020ddpm} and trained on vast datasets~\cite{schuhmann2022laion, bain2021webvid}. Following the scaling recipe seen in NLP and 2D vision~\cite{achiam2023gpt, alayrac2022flamingo, bai2024lvm}, the Large Reconstruction Model (LRM) has been introduced to learn general 3D reconstruction priors~\cite{hong2023lrm}. For object-level reconstruction, LRMs have shown impressive reconstruction quality using either single-view or sparse-view inputs~\cite{hong2023lrm, wang2023pflrm, zhang2024gslrm, jin2024lvsm}, enabling a range of applications~\cite{li2023instant3d, zhang2024vision}.

Despite progress, enhancing LRM for reconstructing \textit{wide-coverage scenes} remains challenging due to two key limitations of training data.
First, scene-level datasets are significantly \textit{smaller in scale} compared to object-level counterparts. For instance, Objaverse~\cite{deitke2023objaverse} contains 800K shape instances, whereas the largest clean scene dataset, DL3DV, includes just 10K scenes. Collecting more intentionally captured scene data is labor-intensive and difficult to scale.
Second, existing scene-level datasets suffer from a \textit{suboptimal data distribution}. They are
often limited by insufficient scene diversity~\cite{dai2017scannet}, small camera motions~\cite{zhou2018stereo, mildenhall2019llff}, noisy content~\cite{tung2024megascenes}, and inaccurate annotations~\cite{MegaDepthLi18}. However, given the inherent complexity of 3D scenes, effective training requires clean and diverse data, especially multi-view images captured by widely spaced cameras with precise camera annotations~\cite{mildenhall2021nerf}.



In this work, we propose scaling up training data for scene-level reconstruction by using \textbf{synthesized data}. Our key idea is to \textbf{eliminate the reliance on semantic information in data generation}, by constructing scenes with non-semantic shape primitives arranged within basic spatial structures. This approach is motivated by our insight that \textit{scene semantics play a minimal role in multi-view reconstruction}, as evidenced by the success of traditional non-semantic methods such as COLMAP~\cite{schonberger2016structure}, MVS~\cite{seitz2006comparison}, NeRF~\cite{mildenhall2021nerf}, and the emerging non-semantic properties of recent feed-forward models~\cite{jiang2023leap, charatan2024pixelsplat, xie2024lrmzero}.
Unlike prior scene generation methods, which aim to replicate real-world scene distributions~\cite{roberts2021hypersim, srivastava2022behavior, wang2020tartanair, raistrick2024infinigen_indoor, fu20213d, zheng2023pointodyssey} and are thus constrained by the complexity of modeling semantics, e.g. object affordances, our approach bypasses these challenges. This simplification enables highly scalable and efficient data generation.

Beyond scalability, synthesized data offers controllability. We control data complexity to facilitate training while loosely aligning it with real-world data distribution to benefit real-world generalization. Through heuristic methods, we regulate key factors, such as geometric complexity, camera pose distribution, materials, and lighting, for creating diverse scenes. Additionally, synthesized data provides precise metadata, such as camera and geometry information, further ensuring improved training stability and effectiveness.


We generate the \textbf{\dataname{}} dataset, comprising \textit{700K} scenes. \dataname{} is \textit{over 50 times larger} than the real dataset DL3DV and significantly scales up training data for LRMs. We utilize \dataname{} to train feed-forward LRMs~\cite{zhang2024gslrm, longlrm} jointly with DL3DV.
Our experiments show a 1.2 to 1.8 dB PSNR gain across diverse test datasets and image resolutions. Moreover, the depth rendering quality is significantly improved, showing a better reconstruction geometry quality. These results highlight the synergy between synthesized and real data. 
Synthesized data excels in scale and provides rich metadata, such as geometry supervision, enabling models to develop a general geometric understanding beyond rendering supervision. Meanwhile, small-scale real data further sharpens the model. 
Interestingly, MegaSynth can also \textit{benefit other 3D tasks}, where a monocular depth estimation model fine-tuned on MegaSynth demonstrates significant improvement.

\section{Related Work}

\noindent\textbf{Scene-level 3D Reconstruction.} Reconstructing scenes has been a long-standing challenge in 3D computer vision. Traditional Structure-from-Motion (SfM) and Multi-view Stereo (MVS) methods, as well as their neural counterparts, adopt a bottom-up approach~\cite{seitz2006comparison, schonberger2016structure, snavely2008modeling, goesele2007multi, furukawa2009accurate, jiang2024forge, smith2024flowmap, wang2024vggsfm}. For instance, COLMAP~\cite{schonberger2016structure} builds from low-level visual cues to more structured geometry through keypoint detection, matching, camera reconstruction, and bundle adjustment.

Learning-based methods encompass both 3D neural scene representations and feed-forward prediction models. Researchers have explored the distinct properties of explicit~\cite{wang2024dust3r, monnier2023differentiable}, implicit~\cite{Sitzmann2020siren, chabra2020deep, mildenhall2021nerf, li2023neuralangelo}, and hybrid 3D representations~\cite{kerbl20233dgs, guedon2024sugar, huang20242dgs, jiang2024cofie} to enhance reconstruction quality, typically optimizing the 3D representation for each scene to demonstrate capability. Meanwhile, generalizable reconstruction models have been developed, where neural networks predict 3D representation attributes in a feed-forward manner. Some approaches follow a traditional bottom-up paradigm, leveraging inductive biases from MVS~\cite{chen2021mvsnerf, yu2021pixelnerf, wang2021ibrnet}, cost volume~\cite{chen2024mvsplat, chen2024mvsplat360}, correspondence cues~\cite{chen2023explicit}, and epipolar geometry~\cite{charatan2024pixelsplat, wewer2024latentsplat, du2023learning}. In contrast, recent work proposes top-down frameworks~\cite{jiang2023leap, wang2023pflrm, wang2024dust3r, leroy2024mast3r} that infer geometry directly and better harness the power of large models. However, some of these works rely on pairwise computations~\cite{wang2024dust3r}, which limits a global understanding of inputs. Our work, in contrast, leverages recent global-aware methods~\cite{zhang2024gslrm, longlrm} and focuses on scaling up training data to advance dense-view reconstruction.

\noindent\textbf{Large Reconstruction Model (LRM).} LRMs have been introduced to scale up generalizable 3D reconstruction methods~\cite{hong2023lrm}, employing scalable network architectures and training on large datasets to learn generic reconstruction priors. Typically, LRMs use Transformers~\cite{vaswani2017transformer, hong2023lrm, jiang2023leap} or U-Nets~\cite{ronneberger2015unet, tang2024lgm} as model backbones, encoding 2D image inputs into 3D representations, e.g., Triplane~\cite{hong2023lrm, wang2023pflrm} and mesh~\cite{wei2024meshlrm, xu2024instantmesh}, enabling high-quality object reconstruction. The following research has focused on enhancing object reconstruction by incorporating generative priors~\cite{xu2023dmv3ddenoisingmultiviewdiffusion, zhang2024relitlrmgenerativerelightableradiance} and designing more scalable training frameworks~\cite{xie2024lrmzero, jiang2024real3d, han2025vfusion3d}. Additionally, novel 3D representations, such as 3D Gaussians~\cite{kerbl20233dgs}, have extended LRMs to scene-level 3D reconstruction~\cite{zhang2024gslrm, longlrm}. However, reconstructing wide-coverage scenes remains challenging due to limited data. To address this, we propose a scalable data generation method considering the non-semantic property of multi-view reconstruction.


\noindent\textbf{Training with Synthesized Data.} 
Leveraging synthesized data for training is essential when available data is insufficient or biased. Synthesized data has been widely applied across fields such as Robotics~\cite{srivastava2022behavior}, Natural Language Processing~\cite{adler2024nemotron}, Computer Vision~\cite{mayer2016large}, and AI for Science~\cite{trinh2024solving}. For example, recent depth estimation methods utilize synthesized data’s accurate ground truth to enhance performance on fine structures~\cite{yang2024depthanythingv2, bochkovskii2024depthpro}. A relevant topic to our work is 3D scene generation, where generated data supports training 3D reconstruction models~\cite{nasiriany2024robocasa, deitke2022️, raistrick2023infinite, yang2021scene, lin2024genusd, nvidia2024edify3d, hollein2023text2room}. However, these methods focus on generating realistic scenes, necessitating semantic accuracy (e.g., object affordance and relationships), which constrains scalability due to the complex procedural rules required for accuracy and diversity. While some recent methods attempt to address this limitation with language models~\cite{yang2024holodeck}, these models often lack spatial awareness and are slow in inference. In contrast, we show that semantics are not essential for multi-view reconstruction, allowing us to create a data generation pipeline free from semantic constraints and capable of generating virtually unlimited training data. 
Previously, non-semantic shape primitives have been used for various object reconstruction and appearance acquisition tasks \cite{xu2018deep,xu2019deep,li2018learning,sang2020single,li2020through}.
Recently, LRM-Zero~\cite{xie2024lrmzero} has used primitive-based methods to generate large-scale data to train large reconstruction models, but it is limited to the object level. 
We focus on more challenging scene-level data synthesis, incorporating control of lighting, object composition, and camera poses for reconstructing wide-coverage scenes from dense-view images. We also present a mixed training framework to leverage the synergy between synthesized and real data. DUST3R~\cite{wang2024dust3r} employs a pre-trained encoder from CroCo~\cite{weinzaepfel2022croco}, which incorporates synthesized data, but its pre-training is limited to 2D image representation learning without directly learning 3D priors. In contrast, our pre-training approach directly targets 3D scenes, enhancing our model’s geometric and texture understanding. We also enable joint training with both synthesized and real data.

\section{Task and Preliminary}
\label{sec:task_and_lrm}
Our goal is to \textit{reconstruct wide-coverage scenes in a feed-forward manner}. Given a set of dense-view images $\{I_i \mid i=1,...,n\}$ with known camera information, the model predicts the attributes of 3D representations. By default, we use $\mathbf{n=32}$ views in our experiments to handle the high complexity of scenes, in contrast to previous sparse-view methods that rely on only 4 to 8 views~\cite{wei2024meshlrm, wang2023pflrm}.

This paper primarily experiments with GS-LRM~\cite{zhang2024gslrm} and Long-LRM~\cite{longlrm}, chosen for their strong reconstruction quality. Both methods predict pixel-aligned 3D Gaussians from posed images with similar model architectures but different backbones: GS-LRM and Long-LRM employ transformer-based and Mamba-based~\cite{gu2023mamba, lieber2024jamba} backbones, respectively.


Given the input views, the models first patchify each image using non-overlapping convolutions, encoding them into feature tokens $\{T_i \mid i=1,...,n\}$ as in ViT~\cite{Dosovitskiy2020vit}. The feature tokens from all images are flattened and concatenated into a feature set, $\mathbf{F}$, which is later processed by the model $\mathcal{M}$. Finally, an MLP decodes Gaussian parameters $\mathbf{G}$ to represent the scene. The process is formulated as follows:
\begin{align}
    \{T_1, \dots, T_n\} &= \{\mathrm{Conv}(I_1), \dots, \mathrm{Conv}(I_n)\}, \\
    \mathbf{F} &= [\mathrm{Flatten}(T_1), \dots, \mathrm{Flatten}(T_n)], \\
    \bar{\mathbf{F}} &= \mathcal{M} (\mathbf{F}), \\
    \mathbf{G} &= \mathrm{MLP}(\bar{\mathbf{F}}),
\end{align}
where $[\cdot, ..., \cdot]$ denotes concatenation, and $\bar{\mathbf{F}}$ represents the updated feature tokens produced by the backbone.

In the next section, we introduce our approach to synthesize data for training these models.
\vspace{-1mm}
\section{Synthesizing the \dataname{} Dataset}
\label{sec: synthesize_data}
\vspace{-1mm}
In this section, we first give an overview of our data synthesis method and then dive deeper to introduce how we control complexity, diversity, and alignment with real data.

\begin{figure*}[htp]
    \centering
    \includegraphics[width=\textwidth]{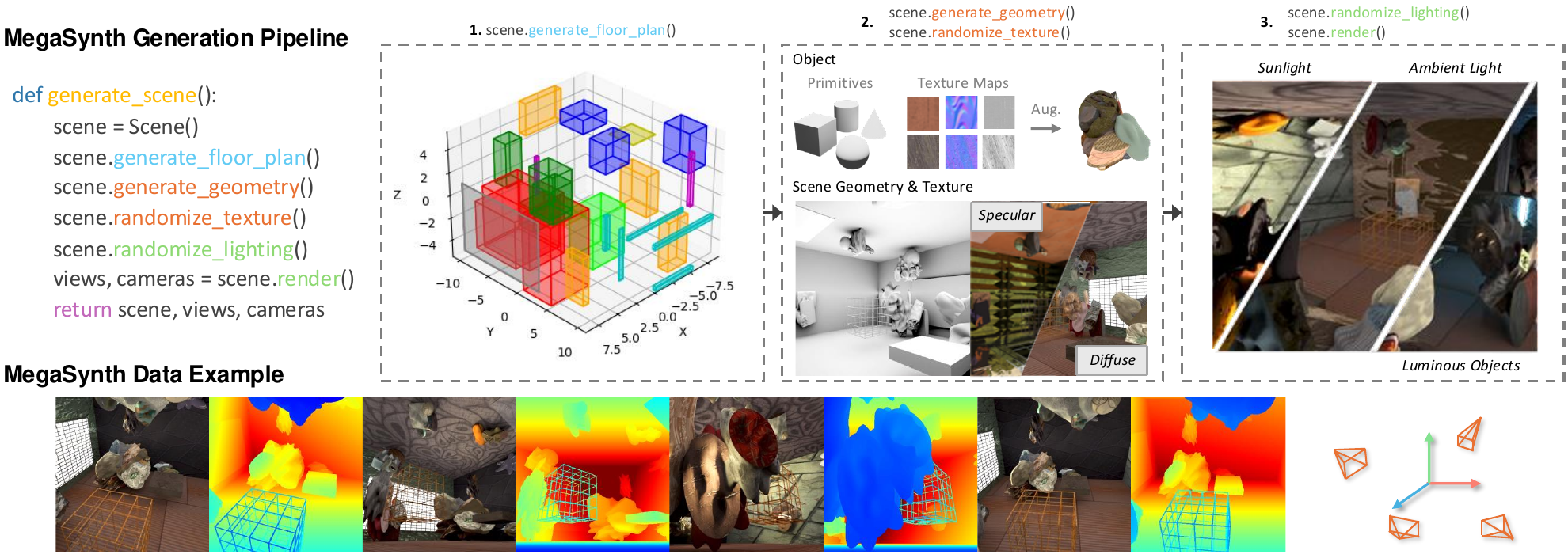}
    \vspace{-0.28in}
    \caption{\small{\textbf{\dataname{} generation pipeline.} We first generate the scene floor plan, where each 3D box represents a shape and different colors represent different object types. We compose shape primitives into objects with geometry augmentations, where these objects further compose the scene. We randomize the texture and lighting, and generate random cameras for rendering.}} 
    \label{fig: data_pipeline}
    \vspace{-0.12in}
\end{figure*}

\vspace{0.5mm}
\noindent\textbf{Overview.} We synthesize \dataname{} using a procedural generation method, as illustrated in Fig.~\ref{fig: data_pipeline}. The process involves: i) generating a scene floor plan, including scene size and object instance box locations, ii) instantiating object geometries with random textures, and iii) randomizing the lighting. During the process, we eliminate high-level scene semantics. We only keep the low-level structural and geometric features of scenes.

\subsection{Scene Floor Plan}
\label{sec:floor_plan}
Without loss of generality, we plan the scene as a cube box and populate it with objects represented by 3D bounding boxes. We randomize the 3D aspect ratio and size of scenes.
We design multiple object box categories to simulate real-world scene geometry structures (visualized as boxes in Fig.~\ref{fig: data_pipeline} with different colors). For example, large object boxes tend to be placed near the ground while small object boxes have more flexible placement options. We parameterize the size, location, and number of each object type, specifying each parameter as a range. This allows us to introduce randomness to improve diversity. Further details of the object box categories and their attribute sampling ranges are provided in the Appendix.

\subsection{Geometry and Texture}
\label{sec:geometry}

The scene floor plan constructed above divides the room space into basic units of object boxes. We then synthesize geometry and assign textures for each geometric shape.

\vspace{0.5mm}
\noindent\textbf{Geometry of general objects.} For each object box, we generate geometry by combining non-semantic shape primitives~\cite{xie2024lrmzero, xu2018deep}, including cubes, spheres, cylinders, and cones. These primitives incorporate diverse geometry patterns, such as flat and curved surfaces, straight and curved lines, and sharp edges. Composing these shapes further increases geometric and topological complexity. Additionally, we apply random height-field augmentations~\cite{xu2018deep} to the primitives, producing surfaces with both concave and convex details.

Different object categories (defined in Sec.~\ref{sec:floor_plan}) utilize varying numbers of shape primitives; for instance, large objects are typically composed of more primitives than small ones, loosely reflecting the complexity distribution of real-world objects. The geometry is instantiated in a canonical space, then rescaled and translated to fit the object box.

\vspace{0.5mm}
\noindent\textbf{Geometry for increasing complexity.} To enhance diversity and alignment with real data, we incorporate two additional types of geometry. 
First, we add thin structures, such as wireframes of shape primitives, enabling the reconstruction of fine-grained geometries.
To further increase diversity, we randomly place solid primitives intersecting with these wireframes. Second, we introduce axis-aligned geometries, such as thin sticks and flat surfaces, to reflect real-world geometry distributions under the Manhattan assumption~\cite{coughlan2000manhattan}.


\vspace{0.5mm}
\noindent\textbf{Texture.} Each shape primitive is assigned a random texture, including a basic color map along with normal, material, and roughness maps. We increase the probability of sampling specular and glass materials, ensuring a closer match to real-world appearances.

\subsection{Lighting}

Real-world images often feature complex lighting conditions. Thus, we design three lighting conditions and randomly compose them to improve the diversity and complexity. Each lighting uses a randomly sampled color and intensity.

\vspace{0.5mm}
\noindent\textbf{Ambient light.} We use the uniformly distributed ambient lighting with a unit brightness by default. The ambient lighting provides consistent illumination across a scene, helping to reveal scene details and stabilizing training.

\vspace{0.5mm}
\noindent\textbf{Sunlight.} Adding sunlight simulates true-to-life lighting effects, making the scene more complex with a higher intensity and casting shadows. We set the sunlight outside of the scene box. To enable the sunlight effect within the scene, we create windows on the walls with random sizes, under the regions that the sunlight covers. To further improve the complexity and diversity, we randomly add window bars implemented as the wireframes and window glasses.

\vspace{0.5mm}
\noindent\textbf{Luminous objects and light bulbs.} 
We randomly turn objects and axis-aligned sticks as lights and place light bulbs in the scene, simulating real-world lighting and increasing diversity. The intensity of object light can be sampled as large values to simulate lighting in dark environments.

\section{Learning 3D Reconstruction on \dataname{}}
\label{sec: training}


In this section, we discuss how we utilize our synthesized \dataname{} (Sec.~\ref{sec: synthesize_data}) to train a feed-forward reconstruction model (i.e., the LRM-based model illustrated in Sec.~\ref{sec:task_and_lrm}).
To reach the goal, we first construct the training data by carefully sampling the camera distribution and rendering the images (Sec.~\ref{sec:render_data}).
We then train our model with a mixed-data training strategy (Sec.~\ref{sec:mixed_training}) with rendering loss and geometry loss (Sec.~\ref{sec:loss_function}).
The details of the training process can be found in the Appendix.






\subsection{Training Data Preparation on \dataname{}}
\label{sec:render_data}

To get training data, we render input views and target supervisions from the synthesized \dataname{} scenes.
We sample cameras and then render RGB and depth images accordingly. 
We do not distinguish the input views and target views, i.e., they will be used interchangeably during training.

The main challenge of this data creation pipeline is the camera pose sampling.
We empirically found that a careful design of camera sampling distribution can largely improve learning efficiency, model generalization, and training stability.
We next detail our camera sampling process.

\vspace{0.5mm}
\noindent\textbf{Basic rules.} 
The cameras are sampled to keep a minimal distance from any objects in the scene, preventing the camera from losing context and avoiding the near-clipping issue. 
We randomly sample the field-of-view (FoV) of cameras, due to the diversity of lenses used in real-world image capture.

\vspace{0.5mm}
\noindent\textbf{Better scene coverage.} 
We heuristically split the scene into the inner and outer spaces, based on the distance to the scene center.
The cameras sampled in outer space always look at the scene center, ensuring better view coverage.
Meanwhile, the cameras in the inner space are encouraged to have more diverse poses, e.g. the orientations are randomly sampled within pre-defined ranges, increasing the diversity and matching real-world camera pose distribution.

\vspace{0.5mm}
\noindent\textbf{Constrained camera baseline.} 
The randomly sampled cameras in the outer part of the scene tend to have large baselines. 
To improve diversity, we choose to sample more scenes and cameras with slightly smaller baselines, aligning with real-world camera distribution.
Thus, instead of sampling camera position in all free space, we first sample a distance range and then sample the camera within the constrained space.





\subsection{Mixed Data Training}
\label{sec:mixed_training}
In training, we leverage distinct advantages from both the synthesized \dataname{} renderings and the real-world dataset (e.g., DL3DV).
The synthesized data, with its diversity and scale, provide a foundation for models to learn general reconstruction priors of geometry, texture, and lighting. Moreover, easy access to accurate metadata (e.g., depth images and noise-free camera information) enhances geometric understanding and stabilizes training.

Meanwhile, real-world data offers authenticity that is hard to synthesize yet crucial for model robustness. For instance, it captures real-world imperfections like sensor noise and lighting artifacts, enhancing the model's robustness for real-world deployment. Additionally, its realistic semantics better align the model with real-world scene distributions.

We find these datasets to be complementary. Our experiments investigate two training strategies to leverage their synergy: (1) pre-training on the large-scale \dataname{} dataset followed by fine-tuning on a smaller real-world dataset; and (2) joint training on both datasets simultaneously. These approaches balance scalability and authenticity.

\subsection{Rendering and Geometry Losses}
\label{sec:loss_function}
We follow the standard method for training large reconstruction models using photometric image rendering losses:
\begin{equation}
    \mathcal{L}_\mathit{img} = \mathrm{MSE}(I_i, \hat{I}_i) + \lambda \cdot \mathrm{Perceptual}(I_i, \hat{I}_i),
    \label{eq: loss}
\end{equation}
where $\lambda$ is the weight for balancing the perceptual loss~\cite{johnson2016perceptual}, $I_i$ is ground-truth target image, and $\hat{I}_i$ is image rendered from predicted 3D Gaussians under target camera poses.

Our synthesized data naturally provides accurate geometry information, which is utilized to supervise the geometry of the 3D Gaussians predicted by the LRM models.
In detail, both GS-LRM and Long-LRM (described in Sec.~\ref{sec:task_and_lrm}) predict pixel-aligned 3D Gaussians, where each Gaussian corresponds to a pixel in the input view. We supervise the center location of the predicted 3D Gaussians using the ground-truth geometry information.
It is formulated as
\begin{equation}
    \mathcal{L}_\mathit{loc} = M \cdot  \mathrm{Smooth\mbox{-}L}_1 (\mathbf{c}, \mathbf{G}_\mathit{loc}),
\end{equation}
where $\mathbf{c}$ and $\mathbf{G}_\mathit{loc}$ are ground-truth and predicted 3D Gaussian location, respectively. The ground-truth Gaussian location $\mathbf{c}$ is computed from the depth maps of input views.
Besides, the loss mask $M$ masks out the pixels where the depth is larger than a threshold (e.g., 100 under the scale-normalized camera coordinate frame).
This mask operation helps avoid numerical instability during training.
This geometry loss proves particularly useful for scene-level reconstruction, which typically involves larger depth ranges, making it challenging to infer geometry solely from photometric cues. Additionally, it enhances the training convergence of the 3D Gaussians, as discussed in Long-LRM.


The final loss function can be formulated as $\mathcal{L}^{S} = \mathcal{L}_\mathit{img}^{S} + \gamma \cdot \mathcal{L}_\mathit{loc}^{S}$, 
where $\gamma$ balances the strength of geometry loss term.

\section{Experiments}

\begin{figure*}[htp]
    \centering
    \includegraphics[width=\textwidth]{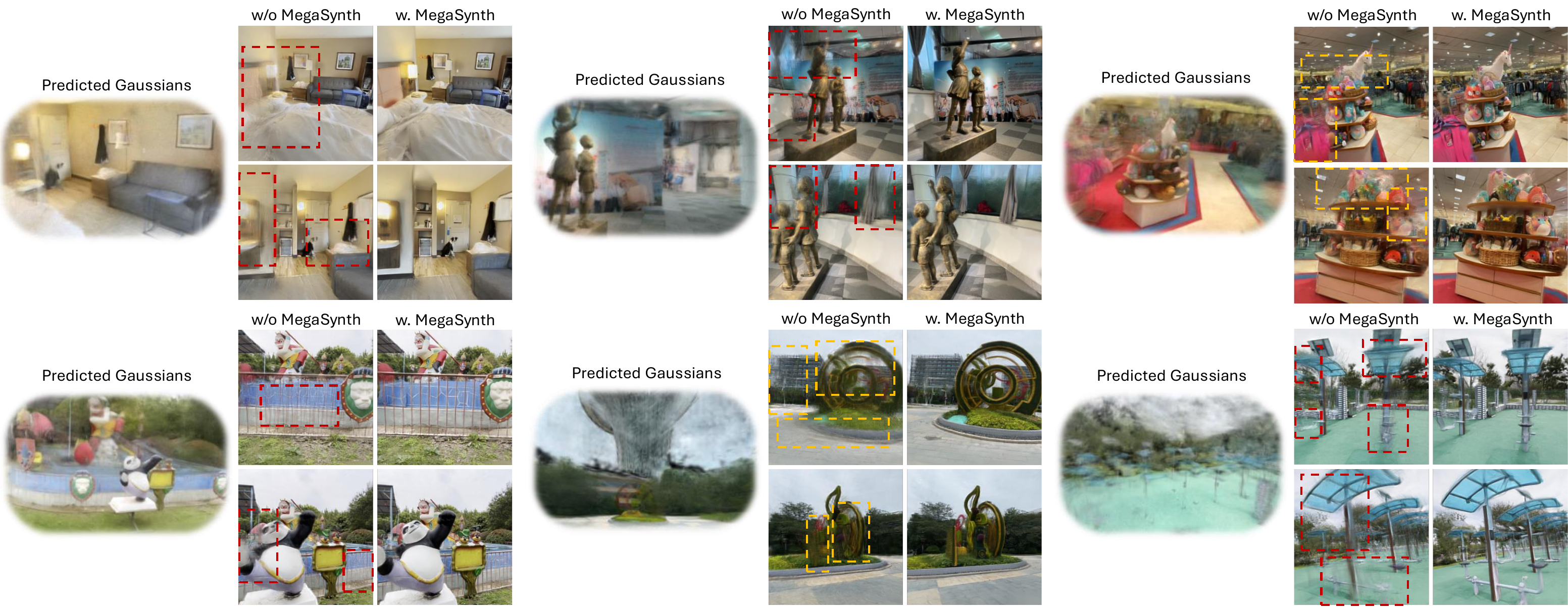}
    \vspace{-0.3in}
    \caption{\small{\textbf{Reconstruction visualization on the in-domain DL3DV data.} 
    The results are from Long-LRM at resolution 256.
    We present both indoor and outdoor results in the first and second rows, respectively. 
    With our \dataname{} (denoted as `w. \dataname{}'), the model performs better on thin structures (e.g., bottom left), complicated lighting (e.g., top middle), and cluttered scenes (e.g., top right).}} 
    \label{fig: vis_dl3dv}
    \vspace{-0.03in}
\end{figure*}

\begin{table*}[t]
\centering
\renewcommand{\arraystretch}{1.25}
\resizebox{1.0\linewidth}{!}{
\begin{tabular}{ll|c|ccc|ccccc|ccc}
\shline
& & \multirow{3}{*}{Inf. Time} & \multicolumn{3}{c|}{\textbf{In-Domain}} & \multicolumn{8}{c}{\textbf{Out-of-Domain} (Zero-shot Generalization)} \\
& &  & \multicolumn{3}{c|}{\textit{DL3DV}} & \multicolumn{5}{c|}{\textit{Hypersim}} & \multicolumn{3}{c}{\textit{MipNeRF360 \& TT}} \\
Model & Training Dataset  & & \multicolumn{1}{c|}{PSNR$_\uparrow$} & \multicolumn{1}{c|}{SSIM$_\uparrow$} & \multicolumn{1}{c|}{LPIPS$_\downarrow$} & \multicolumn{1}{c|}{PSNR$_\uparrow$} & \multicolumn{1}{c|}{SSIM$_\uparrow$} & \multicolumn{1}{c|}{LPIPS$_\downarrow$} & \multicolumn{1}{c|}{AbsRel$_\downarrow$} & \multicolumn{1}{c|}{$\delta_{1\uparrow}$} & \multicolumn{1}{c|}{PSNR$_\uparrow$} & \multicolumn{1}{c|}{SSIM$_\uparrow$} & \multicolumn{1}{c}{LPIPS$_\downarrow$}  \\
\hline \hline
\textsc{Resolution} 128, 32 \textsc{input views} \\
\hline
3DGS~\cite{kerbl20233dgs} & N.A. (Per-scene Optimization) & 5.2 min & 24.27 & 0.817 & 0.166 & 20.67 & 0.672 & 0.293 & 0.320 & 0.715 & 16.46 & 0.458 & 0.405\\
\hline 
Long-LRM~\cite{longlrm} & DL3DV & \multirow{2}{*}{0.12 sec} & 24.18 & 0.812 & 0.173  & 23.41	& 0.790 & 0.210 & 0.272 & 0.763 & 19.68 & 0.569 & 0.312\\
Long-LRM (ours) & DL3DV + \dataname{} & & 25.44 & 0.853 &  0.136 & 25.01 & 0.836 & 0.164 & \textbf{0.258} & 0.792 & 20.86 & 0.652 & 0.249\\
\hline
GS-LRM~\cite{zhang2024gslrm} & DL3DV & \multirow{2}{*}{\textbf{0.11 sec}} & 24.60 & 0.824  & 0.161 & 23.89 & 0.806 & 0.195 & 0.291 & 0.772 & 19.93 & 0.601 & 0.289\\
GS-LRM  (ours) & DL3DV + \dataname{} & & \textbf{25.75} & \textbf{0.859} & \textbf{0.130} & \textbf{25.46} & \textbf{0.846} & \textbf{0.154} & \textbf{0.258} & \textbf{0.800} & \textbf{21.19} & \textbf{0.672} & \textbf{0.235}\\
\hline \hline
\textsc{Resolution} 256, 32 \textsc{input views} \\
\hline
3DGS~\cite{kerbl20233dgs} & N.A. (Per-scene Optimization) & 6.4 min & 23.26 & 0.778 & 0.206  & 21.75 & 0.690 & 0.294 & 0.319 & 0.709 & 16.06 & 0.436 & 0.421\\
\hline
Long-LRM~\cite{longlrm} & DL3DV & \multirow{2}{*}{\textbf{0.35 sec}} & 23.71 & 0.779 &  0.236  & 22.51 &	0.767 & 0.267 & 0.291 & 0.753 & 18.61 & 0.465 & 0.421\\
Long-LRM  (ours) & DL3DV + \dataname{} & & \textbf{25.14} & \textbf{0.828} &  \textbf{0.186} & \textbf{24.26} & \textbf{0.817} & \textbf{0.210} & \textbf{0.255} & \textbf{0.794} & \textbf{19.84} & \textbf{0.555} & \textbf{0.339}\\
\shline
\end{tabular}
}
\vspace{-0.12in}
\caption{\small{\textbf{Evaluation results against baseline methods.} We report results at resolutions of 128 and 256. For resolution 256, we only report results of Long-LRM as transformer-based GS-LRM is too slow. Our models are pre-trained on \dataname{} and then tuned on DL3DV. We report NVS quality on all data and evaluate reconstruction by measuring geometry accuracy (rendered depth accuracy) on Hypersim.}
}
\vspace{-0.2in}
\label{table: main}
\end{table*}

In this section, we describe the experimental setting and present evaluation results. Due to the space limit, implementation and training details are in the Appendix.

\begin{figure*}[htp]
    \centering
    \includegraphics[width=\textwidth]{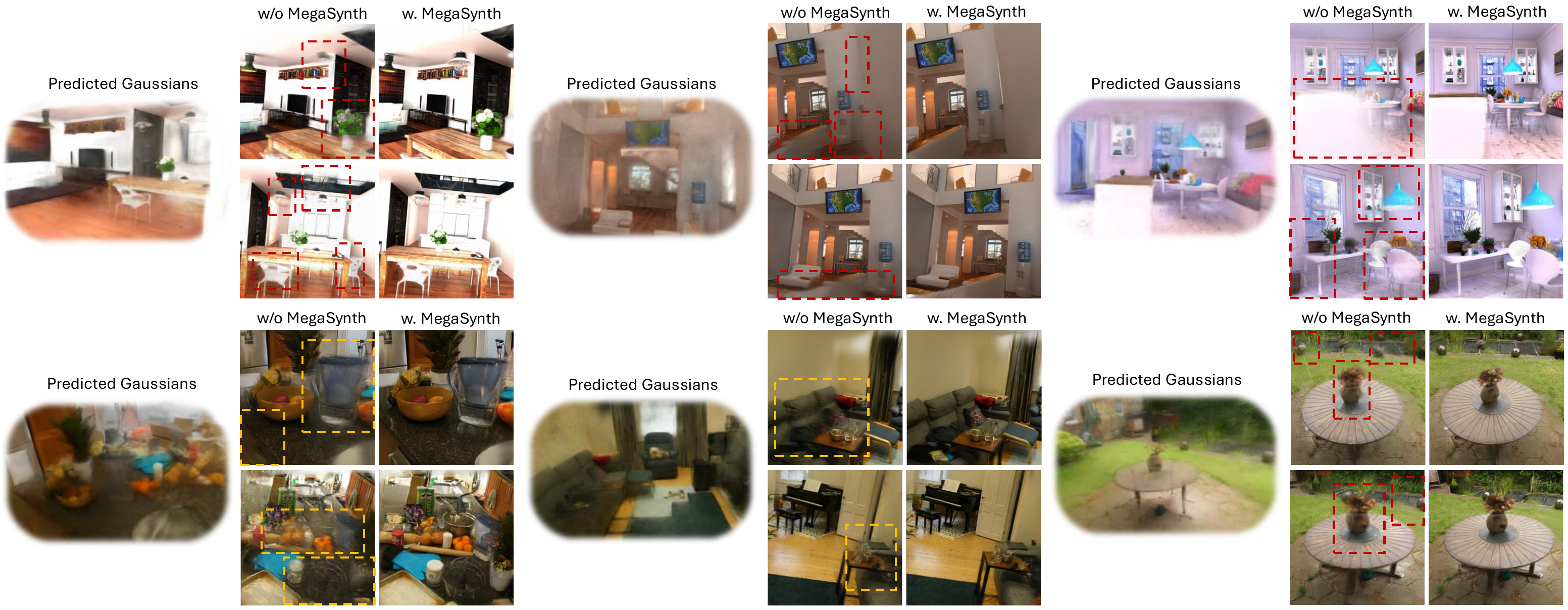}
    \vspace{-0.3in}
    \caption{\small{\textbf{Reconstruction visualization on the out-of-domain data.} The results are from Long-LRM at resolution 256.
    We include results for both 
    Hypersim and MipNeRF360 are presented in the first and second rows, respectively. 
    }} 
    \label{fig: vis_ood}
\end{figure*}

\begin{table*}[t]
    \tiny
    \centering
    \tablestyle{5pt}{1.1}
    \begin{minipage}[t]{0.65\linewidth}
    \tablestyle{5pt}{1.2}
     \centering
    \vspace{-0.05in}
    \resizebox{\linewidth}{!}{
    \begin{tabular}{cccc|c|c|c|c|c|c|c|c}\shline
     & & & & \multicolumn{4}{c|}{\textbf{\dataname{}-only Training }} & \multicolumn{4}{c}{\textbf{Real Data Tuning}} \\

     & Data & \multirow{2}{*}{$\mathcal{L}^S_\text{loc}$} & Scale & \multicolumn{4}{c|}{(Trained w. only \dataname{})} & \multicolumn{4}{c}{(Using DL3DV)} \\
     
     & Control & & Up & Fail. Iter. & \multicolumn{1}{c|}{PSNR$_\uparrow$} & \multicolumn{1}{c|}{SSIM$_\uparrow$} & \multicolumn{1}{c|}{LPIPS$_\downarrow$} & Fail. Iter. & \multicolumn{1}{c|}{PSNR$_\uparrow$} & \multicolumn{1}{c|}{SSIM$_\uparrow$} & \multicolumn{1}{c}{LPIPS$_\downarrow$} 
     \\ \hline\hline
     (0) & \xmark & \xmark & \xmark & 70k & 17.18 & 0.519 & 0.445 &	80k & 18.44 & 0.577 & 0.418\\
     (1) & \cmark & \xmark & \xmark & 45k & 18.71 & 0.601 & 0.384 & 57k & 21.87	& 0.738 & 0.266\\
     (2) & \cmark & \cmark & \xmark & - & 20.72 & 0.691 & 0.300 & - &  25.12 & 0.835 & 0.171\\ 
     (3) & \cmark & \cmark & \cmark & - & \textbf{21.07} & \textbf{0.698} & \textbf{0.292} & - & \textbf{25.46} &	\textbf{0.846} & \textbf{0.154} \\ \shline
    \end{tabular}
    }
    \vspace{-0.12in}
    \caption{\small{\textbf{Ablation study on data control, property and quantity.} Results are reported on the Hypersim dataset with resolution 128. We also report the number of iterations before the job fails. Please see Appendix for data control details for experiment (0). The default data is composed of 100K examples, and the scaled one contains 700K examples. }}
    \vspace{-0.1in}
    \label{table: ablation}
    \end{minipage}
    \hspace{0.2in}
    \begin{minipage}[t]{0.28\linewidth}
    \vspace{-0.05in}
    \captionsetup{type=table}
    \resizebox{\linewidth}{!}{
    \begin{tabular}{l|c|c|c}\shline
     & \multicolumn{3}{c}{\textit{DL3DV-Test-Indoor}} \\
     Data & \multicolumn{1}{c|}{PSNR$_\uparrow$} & \multicolumn{1}{c|}{SSIM$_\uparrow$} & \multicolumn{1}{c}{LPIPS$_\downarrow$}  \\ \hline \hline
     DL3DV & 25.41 & 0.853 & 0.150 	\\
     DL3DV + \dataname{} & \textbf{26.75} & \textbf{0.890} & \textbf{0.116} 
     \\ \hline \hline
     & \multicolumn{3}{c}{\textit{DL3DV-Test-Outdoor}} \\
     Data & \multicolumn{1}{c|}{PSNR$_\uparrow$} & \multicolumn{1}{c|}{SSIM$_\uparrow$} & \multicolumn{1}{c}{LPIPS$_\downarrow$} \\ \hline \hline
     DL3DV & 23.09 & 0.771 & 0.183	\\
     DL3DV + \dataname{} & \textbf{23.89} & \textbf{0.803} & \textbf{0.157} \\
     \shline
    \end{tabular}}
    \vspace{-0.12in}
    \caption{\small{\textbf{Performance gains on indoor and outdoor test data.} 
    Results are from 128-resolution GS-LRM.
    Test data split details are in Sec.~\ref{sec:datasets}.} }
    \vspace{-0.32in}
    \label{table: ablation_indoor_outdoor}
    \end{minipage}
\end{table*}

\vspace{-0.05in}
\subsection{Datasets} 
\label{sec:datasets}
Besides our \dataname{}, we use three datasets in our paper, where DL3DV is the only one we take into training, i.e., others are evaluation-only.


\vspace{0.5mm}
\noindent\textbf{DL3DV}~\cite{ling2024dl3dv}\footnote{We refer to the DL3DV-10K dataset. Only 7K scenes were used in this project as it was completed before the full release.} is a large-scale dataset capturing diverse real-world scenes. 
We split it into $6723$ and $400$ scenes for training and performing evaluation, respectively. 
The $400$ testing scene is composed of the DL3DV benchmark (140 outdoor scenes) and 260 indoor scenes held out from its official training set to balance the indoor-outdoor ratio.

\vspace{0.5mm}
\noindent\textbf{Hypersim}~\cite{roberts2021hypersim} is a synthetic 3D indoor scene dataset with 
ultra photo-realistic renderings, aimed at testing the generalization capability to out-of-distribution indoor scenes. Hypersim is challenging due to its complicated geometry, extreme lighting conditions, and large camera baselines. Hypersim also provides high-quality depth ground truth. We use a test set composed of 302 scenes. 

\vspace{0.5mm}
\noindent\textbf{MipNeRF360, Tanks \& Temples (TT)}~\cite{barron2022mip, knapitsch2017tanks} includes 11 scenes for further testing the out-of-domain generalization capability of models on real data.

\subsection{Evaluation and Baselines}
We use $32$ views as input and use $32$ target novel view images for evaluation. The input and target views are non-overlap and are evenly sampled.
We compare with three baselines:

\noindent\textbf{GS-LRM and Long-LRM trained on DL3DV.} These two baselines aim at validating \textit{the effectiveness of our proposed data.} In detail, we train the GS-LRM and Long-LRM models on the largest real scene-level dataset, DL3DV, using the same training setting as ours.

\vspace{0.3mm}
\noindent\textbf{Optimization-based 3DGS}~\cite{kerbl20233dgs}. This baseline aims at validating the overall performance of our method, as the optimization-based 3DGS usually demonstrates a promising reconstruction quality. We use known camera information to get point cloud initialization from the $32$ input views using COLMAP. We use official training hyper-parameters.

Additionally, we note that comparing with more advanced 3DGS methods is not the focus our work, as our target is scaling up training data for improving feed-forward methods. Our contributions can be directly ablated by comparing with LRMs trained without our data.

\vspace{-0.25mm}
\subsection{Results}
\label{sec:results}
\vspace{-0.25mm}

Table~\ref{table: main} presents our results, demonstrating that training with both DL3DV and our \dataname{} dataset significantly improves model performance, with PSNR gains ranging from 1.2 to 1.8 dB. This improvement is consistent across model architectures (GS-LRM and Long-LRM), testing data (both in-domain DL3DV and out-of-domain datasets), image resolutions, and evaluation metrics, highlighting the effectiveness of our synthesized \dataname{} in enhancing the reconstruction quality of LRMs for wide-coverage scenes. Moreover, the rendering depth quality improves significantly as evaluated on Hypersim, showing the benefit of improving geometry quality by training with \dataname{}. Fig.~\ref{fig: vis_dl3dv} and Fig.~\ref{fig: vis_ood} visually compare the reconstruction results for models with and without \dataname{}.
We observe remarkable improvements in scenes with complicated scene structures, geoemtry, material and lighting, aligning with data generation designs (Sec.~\ref{sec: synthesize_data}). 
Our approach also achieves substantially better results than the optimization-based 3DGS method while offering much faster inference speeds (e.g., over 2000 times speed-up from 5 minutes to 0.1 seconds).

We observe a notable trend of utilizing \dataname{}. 
The performance gains with \dataname{} on out-of-domain data are often larger than those on in-domain data. For example, Long-LRM achieves PSNR gains of 1.6 and 1.8 dB on Hypersim at resolutions of 128 and 256, respectively, surpassing the 1.3 and 1.4 dB improvements observed on the in-domain DL3DV dataset. GS-LRM results exhibits a similar pattern. The results underscore \dataname{}'s effectiveness in enhancing the generalization capability of LRMs.


\subsection{Ablation Studies}
\vspace{-1mm}
In this section, we examine the impact of \dataname{} data quality, quantity, properties, and training paradigms for utilizing synthesized data. Without additional specification, the default experimental setup is the resolution-128 GS-LRM with pre-training + fine-tuning training protocol.

\vspace{0.5mm}
\noindent\textbf{\dataname{} data quality, quantity and property.} Table~\ref{table: ablation} presents our results.
In general, we observe a positive correlation between performance of \dataname{}-only training and subsequent real-data fine-tuning, underscoring the value of \dataname{} in model training. Specifically, we refer \dataname{}-only training to the model trained after the per-training stage using only \dataname{}.

In Table~\ref{table: ablation} (0), training with a basic version of \dataname{} without controlling the data diversity and complexity results in lower performance than training with real data alone (Table~\ref{table: ablation_training_paradigm}), suggesting that unregulated synthesized data fails to enhance training. Additionally, we observe training instability, with pre-training and fine-tuning failing after around 70K iterations. We hypothesize that the high data randomness contributes to this instability, impeding effective learning and negatively affecting fine-tuning of real data.

Introducing control of data distribution, as shown in Table~\ref{table: ablation} (1), improves both \dataname{}-only training and real-data fine-tuning performance, emphasizing the importance of data quality and effectiveness of our data control method. However, training instability worsens, likely due to the increased complexity that amplifies training challenges.

Incorporating metadata during training mitigates this instability. Table~\ref{table: ablation} (2) shows that adding geometrical supervision, $\mathcal{L}_{\text{loc}}^S$, significantly improves stability and overall performance. This result underscores a key advantage of \dataname{}: the ability to provide additional ground-truth data. Expanding the dataset to include more scenes (i.e., 700K scenes in total), as in Table~\ref{table: ablation} (3), yields additional gains, showing the benefit of scale.


\begin{table*}[t]
    \tiny
    \centering
    \tablestyle{5pt}{1.1}
    \begin{minipage}[t]{0.3\linewidth}
    \captionsetup{type=table}
    \vspace{0.1in}
    \resizebox{\linewidth}{!}{
    \begin{tabular}{l|c|c|c}\shline
     & \multicolumn{3}{c}{\textit{Hypersim}} \\
      & 
     \multicolumn{1}{c|}{PSNR$_\uparrow$} & \multicolumn{1}{c|}{SSIM$_\uparrow$} & \multicolumn{1}{c}{LPIPS$_\downarrow$}  \\ \hline \hline
     DL3DV Real-only & 23.89 & 0.806 &	0.195	 \\
     \dataname{}-only & 21.50 & 0.719 & 0.272\\
     \hline
     Joint Training & 25.33 & 0.844 & 0.157 \\
     Pre-training + Fine-tuning & \textbf{25.46} & \textbf{0.846} & \textbf{0.154}\\ \shline
    \end{tabular}}
    \vspace{-0.1in}
    \caption{\textbf{Ablation study on the training framework to leverage \dataname{}.} Results are reported with GS-LRM.}
    \label{table: ablation_training_paradigm}
    \end{minipage}
    \hspace{0.15in}
    \begin{minipage}[t]{0.32\linewidth}
     \centering
     \captionsetup{type=figure}
    \includegraphics[width=\textwidth]{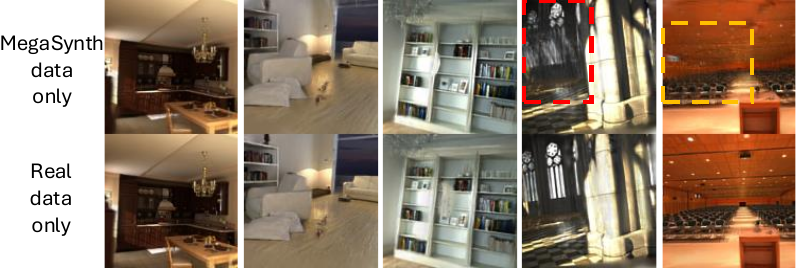}
    \vspace{-0.23in}
    \caption{\textbf{Visual comparison between training with only \dataname{} and only real data.} We include two failure cases of \dataname{}-only with failures highlighted.
    }
    \label{fig: ablation_syn_only}
    \end{minipage}
    \hspace{0.15in}
    \begin{minipage}[t]{0.26\linewidth}
     \centering
     \captionsetup{type=figure}
     \vspace{-0.1in}
    \includegraphics[width=\textwidth]{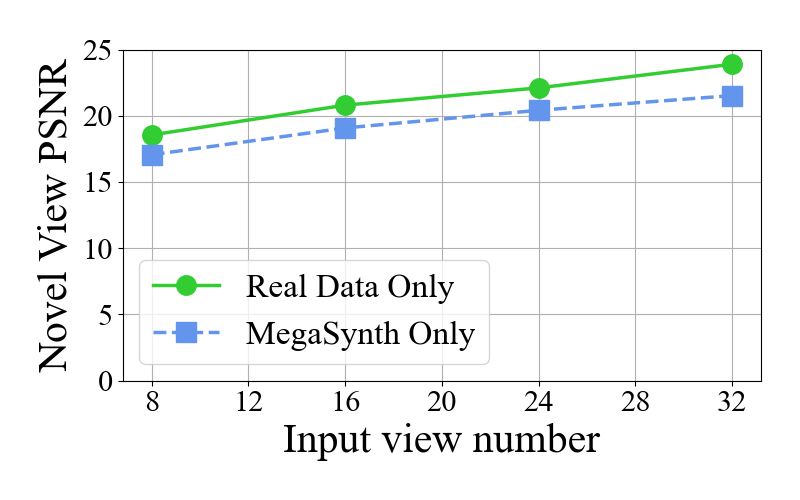}
    \vspace{-0.35in}
    \caption{\textbf{Analysis of real data only and \dataname{}-only performance} with different number of input views.
    }
    \label{fig: analysis_curve}
    \end{minipage}
    \vspace{-0.15in}
\end{table*}

\vspace{0.5mm}
\noindent\textbf{Indoor and outdoor improvements.} 
We analyze the performance gains in Table~\ref{table: ablation_indoor_outdoor}, focusing on both indoor and outdoor test data from DL3DV. 
Although our synthesized \dataname{} data is primarily focusing on indoor scenes, we observe improvements in outdoor scenes as well, with a notably larger performance gain on indoor scenes. 
This suggests that the \dataname{} contributes to a generalized enhancement in geometric and appearance understanding, enabling broader generalization across diverse environments.
At the same time, improving \dataname{} with outdoor characteristics would be an interesting direction.

\vspace{0.5mm}
\noindent\textbf{Training strategies.} We evaluate different strategies for utilizing \dataname{}. As shown in Table~\ref{table: ablation_training_paradigm}, training exclusively on \dataname{} (row 2) achieves performance comparable to training on real data (row 1), highlighting the effectiveness of \dataname{} and supporting our hypothesis that explicit semantics are not required for training scene reconstruction models. 
We visualize the results in Fig.~\ref{fig: ablation_syn_only}.
We find the model performs closely in most of the scenes but is much worse on complicated geometry patterns and large scene scales that are hard to model in synthesized data.

We further compare two approaches: (i) joint training on both synthesized \dataname{} and real data in row 3, and (ii) pre-training on \dataname{} followed by fine-tuning on real data in row 4. As shown in Table~\ref{table: ablation_training_paradigm}, the second approach yields slightly better performance, though the performance gap is minimal. This suggests that the model effectively learns the joint distribution of synthesized and real data without catastrophic forgetting during fine-tuning, indicating a degree of distribution alignment between \dataname{} and real data. Additionally, this experiment confirms that the performance gain results from the enhanced reconstruction capability acquired through \dataname{}, rather than simply from additional training iterations. 

\subsection{Analysis}


We perform a more detailed analysis of \dataname{}, especially its effectiveness against other synthetic data and application to other 3D tasks.


\vspace{0.5mm}
\noindent\textbf{Analysis on different numbers of input views.} We extend our model trained with \dataname{} to scenarios with fewer input views, training GS-LRM with inputs of 8, 16, 24, and 32 views using either real-world data alone or a combination of real-world and \dataname{} data. As shown in Fig.~\ref{fig: analysis_curve}, GS-LRM trained with both DL3DV and synthesized \dataname{} data demonstrates improved performance as the number of input views increases. Notably, an almost constant performance gap remains regardless of the number of views, which we attribute to the semantic gap between DL3DV and \dataname{}. These results highlight the effectiveness of \dataname{} for sparse-view reconstruction and suggest that semantic alignment is not a primary driver of 3D reconstruction performance.

\begin{table}[t]
    \centering
    \small
    \begin{minipage}[t]{\linewidth}
        \resizebox{\linewidth}{!}{%
        \begin{tabular}{l|c|c|c|c|c|c|c|c|c} \shline
            & \multicolumn{3}{c|}{\textit{DL3DV}} & \multicolumn{3}{c|}{\textit{Hypersim}} &\multicolumn{3}{c}{\textit{MipNeRF360 \& TT}} \\ 
             & \multicolumn{1}{c|}{PSNR$_\uparrow$} & \multicolumn{1}{c|}{SSIM$_\uparrow$} & \multicolumn{1}{c|}{LPIPS$_\downarrow$}  & \multicolumn{1}{c|}{PSNR$_\uparrow$} & \multicolumn{1}{c|}{SSIM$_\uparrow$} & \multicolumn{1}{c|}{LPIPS$_\downarrow$}  & \multicolumn{1}{c|}{PSNR$_\uparrow$} & \multicolumn{1}{c|}{SSIM$_\uparrow$} & \multicolumn{1}{c}{LPIPS$_\downarrow$} 
             \\ \hline \hline
            DL3DV & 18.31 & 0.555 & 0.391 & 18.43 & 0.602 & 0.373 & 15.59 & 0.550 & 0.332\\ 
            DL3DV+Kubric~\cite{greff2022kubric} & 18.28 & 0.552 & 0.395 & 18.46 & 0.600 & 0.375 & 15.49 & 0.546 & 0.340\\
            DL3DV+Front3D~\cite{fu20213d} & 18.40 & 0.558 & 0.389 & 18.48 & 0.603 & 0.370 & 15.63 & 0.551 & 0.329\\ 
            DL3DV+\dataname{} & \textbf{19.58} & \textbf{0.592} & \textbf{0.338} & \textbf{19.88} & \textbf{0.638} & \textbf{0.324} & \textbf{16.72} & \textbf{0.592} & \textbf{0.303} \\ \shline
        \end{tabular}%
        }
        \vspace{-0.12in}
        \caption{\textbf{Comparison with other synthetic datasets.} We report results with 8 input views and GS-LRM under resolution 128.}
        \label{tab: analysis_other_syn_data} 
    \end{minipage}%
    \vspace{-0.08in}
\end{table}

\begin{table}[t]
    \centering
    \small
    \begin{minipage}[t]{0.47\linewidth}
        \centering
        \resizebox{0.9\linewidth}{!}{%
        \begin{tabular}{c|c|c} \hline
            &  AbsRel ($\downarrow$) & $\delta_{1} (\uparrow)$ \\ \hline \hline
            Depth Anything V2 & 0.213 & 0.761\\
            Tuned on MegaSynth & \textbf{0.158} & \textbf{0.799} \\ \hline
        \end{tabular}%
        }
        \vspace{-0.1in}
        \caption{\small{\textbf{\dataname{} benefits monocular depth estimation. }}}
        \label{tab: analysis_depth_anything}
    \end{minipage}%
    \hfill 
    \begin{minipage}[t]{0.48\linewidth}
        \centering
        \resizebox{0.99\linewidth}{!}{%
        \begin{tabular}{l|c|c|c} \hline
            & DL3DV & Ours & Front3D \\ \hline \hline
            Geom. Difficulty ($\downarrow$) & 1.65 & \textbf{1.35} & 3.00 \\ 
            Diversity ($\downarrow$) & \textbf{1.40} & 1.60 & 3.00\\ \hline
        \end{tabular}%
        }
        \vspace{-0.12in}
        \caption{\small{\textbf{User study of data difficulty and diversity.}}}
        \label{tab: r6}
    \end{minipage}
    \vspace{-0.22in}
\end{table}

\vspace{0.5mm}
\noindent\textbf{Advantages over other synthetic datasets.} We experiment with using other synthetic datasets for training LRMs. As shown in Table~\ref{tab: analysis_other_syn_data}, Kurbic~\cite{greff2022kubric} (data released in SRT~\cite{sajjadi2022srt}) and Front3D~\cite{fu20213d} fail to improve LRM performance, while \dataname{} benefits the model across all test datasets consistently. In detail, Kurbic contains 1 million scenes radnomly composed by realistic 3D assets; Front3D is composed of 6,000 indoor scenes designed by artists. The results imply that realistic 3D assets or scene composition is not the gaurantee for improving reconstruction quality. Instead, reconstruction model benefits from data with better non-semantic quality, e.g. geometry difficulty and scene diversity.


\vspace{0.5mm}
\noindent\textbf{\dataname{} also helps other tasks.} We fine-tune Depth Anything V2 ViT-B model on \dataname{} and evaluate on Hypersim. Results in Table~\ref{tab: analysis_depth_anything} shows that \dataname{} helps improving monocular depth estimation, demonstrating the potential of \dataname{} to be used for other 3D tasks.

\vspace{0.5mm}
\noindent\textbf{Comparison with real data.} Tab.~\ref{tab: r6} presents a user study ranking geometry difficulty and scene diversity of datasets, showing our comparability with real data and advantage over the other synthetic data Front3D. Please see more analysis on measuring alignment with real data in Appendix.

\section{Conclusion}
\vspace{-0.5mm}
We introduce \dataname{}, a non-semantic procedurally generated dataset, to improve LRMs for reconstructing wide-coverage scenes. \dataname{} benefits from its scalability and controllability, improving the model's understanding of geometry and appearance. Experiments show \dataname{}'s capability of improving LRM reconstruction quality via both pre-training and joint training. The performance gains are consistent over different model architectures, test data domains, and input/output resolutions. Interestingly, LRMs trained sorely with \dataname{} demonstrate comparable performance with using real data, demonstrating that reconstruction is almost a non-semantic/low-level task.

\section*{Acknowledgements}
The work was done while Hanwen Jiang, Desai Xie, Ziwen Chen, and Haian Jin were interns at Adobe Research.
We thank Kalyan Sunkavalli for the support and feedback. 
Qixing Huang would like to acknowledge NSF IIS 2047677 and NSF IIS 2413161.

{\small
\bibliographystyle{ieee_fullname}
\bibliography{egbib}
}

\appendix
\newpage

\section{\dataname{} Details}
\label{appendix: data_details}
In this section, we include more details of our \dataname{} generation method. We introduce the details according to the sections in the main paper, i.e. scene floor plan, geometry and texture, and lighting.

\subsection{Scene Floor Plan}

We define the parameters of the scene size and object box in Table~\ref{tab: scene_floor plan_1} and Table~\ref{tab: scene_floor plan_2}, including the categories, types, size ranges, height ranges, and probabilities. These object boxes are placed randomly in the scene, except for some categories, i.e. on-ground small box, on-roof box, and on-wall box, which have pre-defined location priors.

\begin{table}[!h]
    \centering
    \renewcommand{\arraystretch}{1.25}
    \resizebox{0.85\linewidth}{!}{
    \begin{tabular}{cc|c}
        \shline
         Scene parameters & Size range &  [17.0, 30.0]\\
         & Height range &  [10.0, 15.0]\\
         \hline
         Object box parameters & \# Categories & 7 \\
         \hline
         Large object box & Size range & [4.0, 8.0] \\
         & Number range & [2, 5] \\
         \hline
         Small object box & Size range & [2.0, 4.0] \\
         & Number range & [4, 8] \\
         & Type 1 & On-ground \\
         & Prob. 1 & 0.5 \\
         & Height range 1 & [2.0, 6.0] \\
         & Type 2 & Atop large box \\
         & Prob. 2 & 0.5 \\
         & Height range 2 & [2.0, 4.0] \\
         \hline
         On-roof object box & Size range & [2.0, 5.0]\\
         & Number range & [2, 4]\\
         & Type 1 & Thin \\
         & Prob. 1 & 0.5 \\
         & Height range 1 & [0.5, 1.5] \\
         & Type 2 & Thick \\
         & Prob. 2 & 0.5 \\
         & Height range 2 & [2, 4] \\
         
        \shline
    \end{tabular}
    }
    \caption{Scene floor plan details part 1.}
    \label{tab: scene_floor plan_1}
\end{table}

\begin{table}[!h]
    \centering
    \fontsize{6pt}{7pt}\selectfont
    \renewcommand{\arraystretch}{1.25}
    \resizebox{0.8\linewidth}{!}{
    \begin{tabular}{cc|c}
        \shline
         On-wall object box & Size range & [2.0, 5.0]\\
         & Number range & [3, 6]\\
         & Type 1 & Thin \\
         & Prob. 1 & 0.5 \\
         & Height range 1 & [0.5, 1.5] \\
         & Type 2 & Thick \\
         & Prob. 2 & 0.5 \\
         & Height range 2 & [2, 4] \\
        \hline
        Wire-frame box & Size range & [3.0, 6.0] \\
        & Number range & [1, 3] \\
        & Height range & [3.0, 6.0] \\
        & Prob. & 0.8 \\
        \hline
        Thin stick box & Length range & [3.4, 18] \\
        & Type 1 & On-wall \\
        & Prob. 1 & 1.0 \\
        & Size 1 & [0.1, 0.6] \\
        & Number 1 & [5, 16] \\
        & Type 2 & In-space \\
        & Prob. 2 & 0.5 \\
        & Size 2 & [0.8, 1.8] \\
        & Number 2 & [2, 6] \\
        \hline
        Axis-aligned box & Size range & [2.0, 5.0] \\
        & Number range & [1, 2] \\
        & Prob. & 0.7 \\
        & Height range & [0.2, 1.0] \\
        \shline
    \end{tabular}
    }
    \caption{Scene floor plan details part 2.}
    \label{tab: scene_floor plan_2}
\end{table}

\subsection{Geometry and Texture.}
We include the details of object geometry in Table~\ref{tab: geometry}. In detail, the probability of using cube, sphere, cylinder and cone primitives are all 0.25 for large, small, on-wall, on-roof and wireframe object. For thin stick and axis-aligned objects, we only use cubes and cylinders. Beside, fro wireframe objects, we use cube, cylinder and torus, where torus has genus, increasing the geometry and topological complexity an diversity. We apply the height field augmentations to all shape primitives except for thin sticks and axis-aligned objects.

\begin{table}[!h]
    \centering
    \fontsize{6pt}{7pt}\selectfont
    \renewcommand{\arraystretch}{1.25}
    \resizebox{1.0\linewidth}{!}{
    \begin{tabular}{cc|c}
        \shline
         Large object & Number of shape primitives & [4, 5, 6, 7, 8, 9] \\
         & Prob. of Number of shape primitives & [0.147, 0.206, 0.294, 0.206, 0.147] \\
         & Primitive types & Default \\
         \hline
         Small object & Number of shape primitives & [2, 3, 4, 5] \\
         & Prob. of Number of shape primitives & [0.25,0.375,0.25,0.125] \\
         & Primitive types & Default \\
         \hline
         On-wall object & Number of shape primitives & [2, 3, 4, 5] \\
         & Prob. of Number of shape primitives & [0.25,0.375,0.25,0.125] \\
         & Primitive types & Default \\
         \hline
         On-roof object & Number of shape primitives & [2, 3, 4, 5] \\
         & Prob. of Number of shape primitives & [0.25,0.375,0.25,0.125] \\
         & Primitive types & Default \\
         \hline
         Wireframe object & Number of shape primitives & [1, 2, 3] \\
         & Prob. of Number of shape primitives & [0.5, 0.25, 0.25] \\
         & Wireframe Primitive types & Torus, Cube, Sphere \\
         & Wireframe thickness & [mean scale/30, mean scale/20] \\
         & Sphere wireframes segments & 8\\
         & Sphere wireframes ring count & 8\\
         & Cube wireframes subdivision & [1, 2, 3] \\
         & Cube wireframes subdivision prob. & [0.33, 0.33, 0.33] \\
         & Torus wireframes minor radius & 0.3 $\cdot$ mean scale \\
         & Torus wireframes major segments & 8 \\
         & Torus wireframes minor segments & 8 \\
         & ProB. Adding intersecting obj. & 0.5 \\
         & Types of intersecting obj. & Default \\
         \hline
         Thin stick object & Number of shape primitives & 1\\
         & Primitive types & Cube or Cylinder \\
         \hline
         Axis-aligned object & Number of shape primitives & 1\\
         & Primitive type & Cube \\
        \shline
    \end{tabular}
    }
    \caption{Object geometry details. 'mean scale' is the average of the geometry size over the three axis.}
    \label{tab: geometry}
\end{table}

We include the details of object textures in Table~\ref{tab: texture}. After we randomly select textures and materials for all instantiated geometry primitives, we randomize the materials to improve complexity and diversity. We also have special deigns for materials of axis-aligned objects. We include details in Table~\ref{tab: texture_aa}.

\begin{table}[!h]
    \centering
    \fontsize{6pt}{7pt}\selectfont
    \renewcommand{\arraystretch}{1.25}
    \resizebox{0.6\linewidth}{!}{
    \begin{tabular}{c|c}
        \shline
         Prob. modify mat. & 0.5 \\
         Prob. modify mat. of slot & 0.4 \\
         Prob. specular scene & 0.2 \\
         Basic roughness range & [0.001, 0.2] \\
         Basic metallic range & [0.001, 1.0] \\
         Specular roughness range & [0.0, 0.05] \\
         Specular metallic range & [0.6, 1.0] \\
        \shline
    \end{tabular}
    }
    \caption{Material details 1.}
    \label{tab: texture}
\end{table}

\begin{table}[!h]
    \centering
    \fontsize{6pt}{7pt}\selectfont
    \renewcommand{\arraystretch}{1.25}
    \resizebox{0.7\linewidth}{!}{
    \begin{tabular}{c|c}
        \shline
        Glass IOR range & [1.4, 1.6] \\
        Glass roughness range  & [0.001, 0.1]\\ 
        Prob. Axis-aligned object glass & 0.8\\
        \shline
    \end{tabular}
    }
    \caption{Material details 2.}
    \label{tab: texture_aa}
\end{table}

\subsection{Lighting Details}
We include the lighting details of sunlight in Table~\ref{tab: lighting_sun}. We include details of luminous objects and light bulbs in Table~\ref{tab: lighting_luminous_objects}.

\begin{table}[!h]
    \centering
    \fontsize{6pt}{7pt}\selectfont
    \renewcommand{\arraystretch}{1.25}
    \resizebox{0.5\linewidth}{!}{
    \begin{tabular}{c|c}
        \shline
        Prob. sunlight & 0.6 \\
        Sunlight strength & [0.2, 2.0]\\ 
        Prob. window glass & 0.5 \\
        Prob. window bar & 0.5 \\
        
        \shline
    \end{tabular}
    }
    \caption{Sunlight details.}
    \label{tab: lighting_sun}
\end{table}

\begin{table}[!h]
    \centering
    \fontsize{6pt}{7pt}\selectfont
    \renewcommand{\arraystretch}{1.25}
    \resizebox{0.8\linewidth}{!}{
    \begin{tabular}{cc|c}
        \shline
        Luminous objects & Applied objects & Thin sticks \\
        & Prob. & 0.7 \\
        & Prob. slot & 0.2 \\
        & Strength range 1 & [0.2, 2.0] \\
        & Prob. strength range 1 & 0.9 \\
        & Strength range 2 & [5.0, 8.0] \\
        & Prob. strength range 2 & 0.1 \\
        \hline
        Light bulb & Num. range & [2, 5] \\
        & Strength range 1 & [0.2, 2.0] \\
        & Prob. strength range 1 & 0.9 \\
        & Strength range 2 & [5.0, 8.0] \\
        & Prob. strength range 2 & 0.1 \\
        \shline
    \end{tabular}
    }
    \caption{Luminous objects and light bulbs details.}
    \label{tab: lighting_luminous_objects}
\end{table}

\section{Model and Training Details}
\label{appendix: training_details}

We include more model and training details as follows.

\noindent\textbf{Training input and target view sampling.} For each training sample in a batch, we randomly sample input views and target views from a pool of 48 views following LRM training strategy~\cite{hong2023lrm}.
The number of input views is always 32.
The number of target views is 12 for 128-resolution experiments, and 8 for 256-resolution experiments to balance the compute cost. We allow the overlap between input and target views during training.
On the \dataname{} dataset, the set of 48 views are randomly sampled.
On the real training data, we evenly sample frames within a distance range, which is sampled from the range of 64 to 128.

\noindent\textbf{Camera pose normalization.} The cameras of the input views are normalized with a random global scale between $1.1$ and $1.6$. For Gaussian rendering, we clip the predicted Gaussian scale of $0.135$. We set a near plane of the Gaussian renderer as $0.1$.

\noindent\textbf{Learning rate and scheduler.} 
In the pre-training stage, we use a peak learning rate of $4e-4$. In the tuning stage using real-world data, we use a smaller peak learning rate of $1e-4$. 
For joint training or training exclusively on real data, we use a learning rate of $4e-4$. 
All experiments adopt a warm-up of $3000$ iterations and cosine learning rate decay.

\noindent\textbf{Batch size.}
For both $128\times 128$ and $256\times 256$ resolution training, we use a batch size of 4 per GPU. 
The experiments are launched on 64 A100 GPUs thus the global batch size is 256.

\noindent\textbf{Training iterations.}
The training iterations for Res-128 and Res-256 are 120K and 80K for each training stage (i.e., pre-training and fine-tuning stages, as well as joint-training), respectively. The final learning rate is decreased to 0 at the end of training. Specifically, we end the pre-training stage at 75K and 55K iterations for experiments on resolution 128 and 256, respectively. Thus, The effective learning rate at the end of pre-training stage is around $1e-4$. The reason is we observe that training with more iterations, especially with a learning rate smaller than $1e-4$, leads to overfit on \dataname{} and makes the fine-tuning stage fail.

\noindent\textbf{Training time cost.} It takes GS-LRM $7$ days for pre-training and fine-tuning, and it takes $4$ days for joint-training, under resolution $128\times 128$.
It takes $11$ days for pre-training and fine-tuning, and it takes $6$ days for joint training on resolution $256\times 256$.  

\noindent\textbf{Gaussian Settings.} We use spherical harmonics of 3 for 3D Gaussians.
We follow all other training hyper-parameters as the original GS-LRM~\cite{zhang2024gslrm} and Long-LRM~\cite{longlrm}. 

\noindent\textbf{Loss weights.}
We set the weights of point location loss (on synthetic data) and perceptual loss as 0.4 and 0.2, respectively. For joint training, we set the probability of sampling data from real and synthetic data as the same. For ablations, we run experiments with resolution $128\times 128$ using GS-LRM.

\noindent\textbf{Training view rendering settings.}
For \dataname{} rendering, we sample 36 and 12 cameras in the outer and inner parts of the scenes, respectively. 
We sample the FoV of cameras within the range of 45 to 70 degrees.

\noindent\textbf{Other details.} In our ablation, quality control means we only use four basic object types without wireframes, think structure and axis-aligned object, without material randomization, and using only ambient lighting.

\section{More Results}
\begin{figure*}
    \centering
    \includegraphics[width=\linewidth]{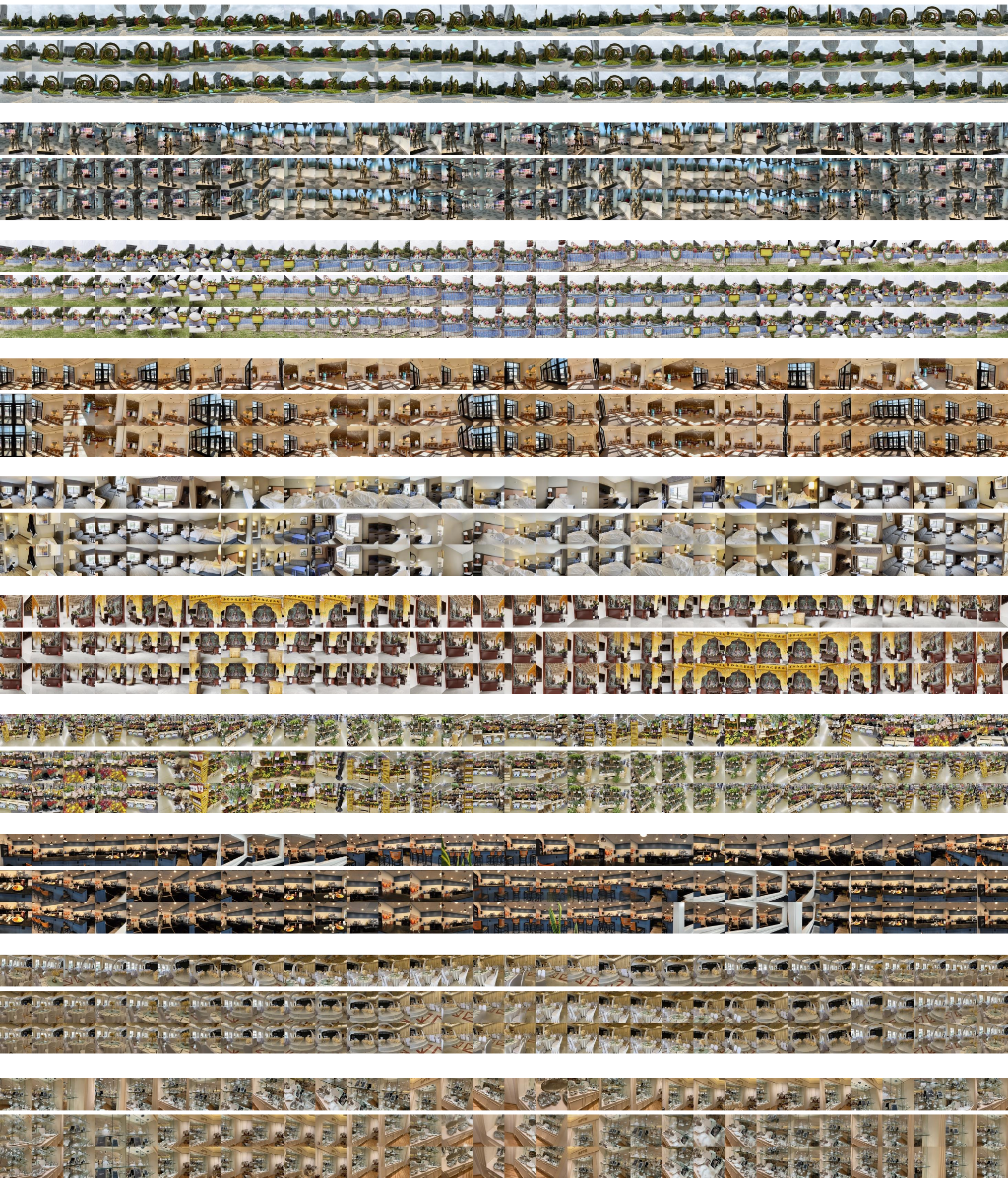}
    \caption{Visualizaton of input views (first row of each example), render target view and ground-truth target views (last two rows of each example). We include results on the DL3DV benchmark data.}
    \label{fig: vis_more}
\end{figure*}

\begin{figure*}
    \centering
    \includegraphics[width=\linewidth]{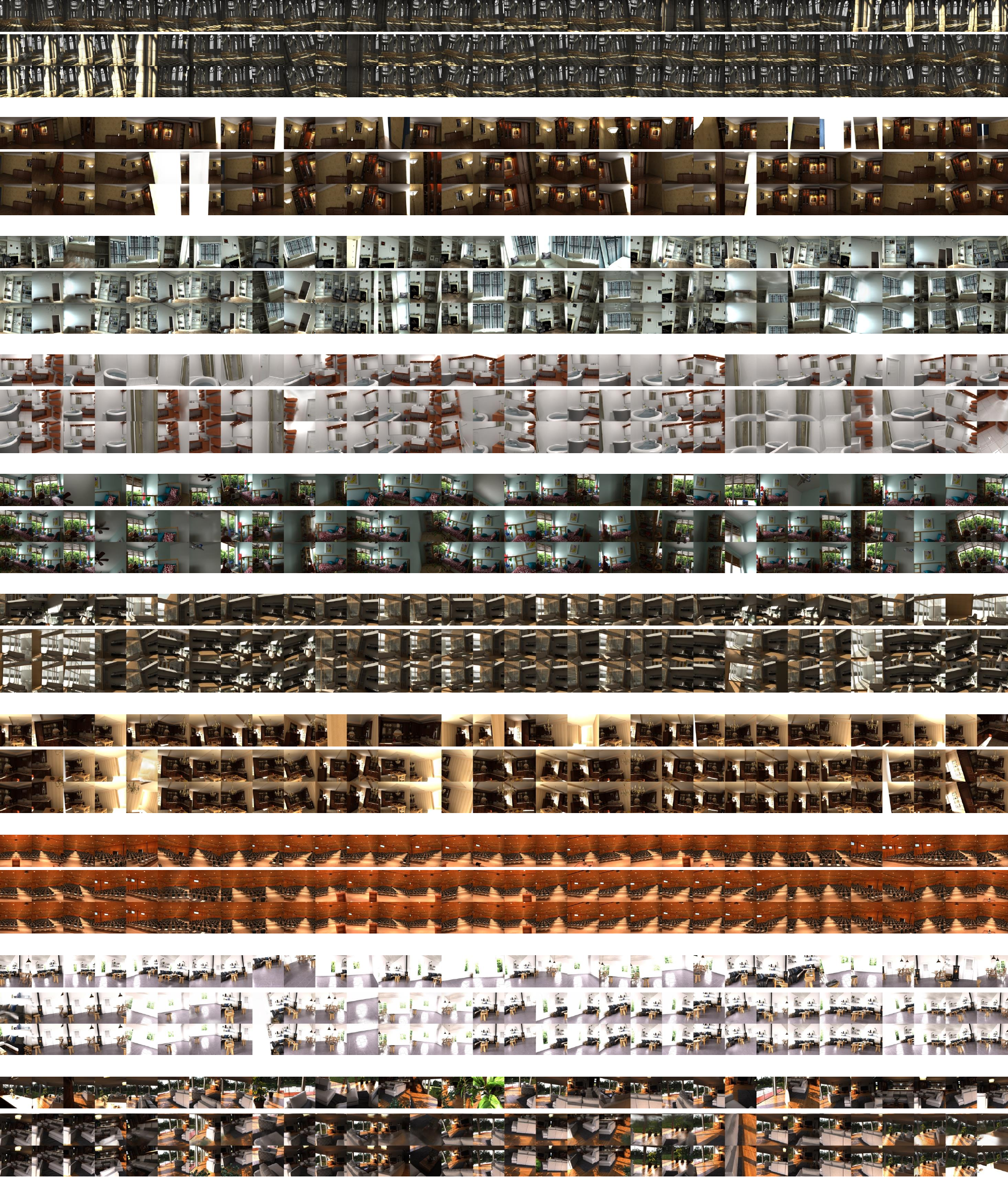}
    \caption{Visualizaton of input views (first row of each example), render target view and ground-truth target views (last two rows of each example). We include results on the Hypersim data.}
    \label{fig: vis_more_2}
\end{figure*}

We include more visualization results with 32 input views and 32 rendered target views as well as the ground-truth target views in Fig.~\ref{fig: vis_more} and Fig.~\ref{fig: vis_more_2}.

\end{document}